\documentclass[a4paper,fleqn]{cas-dc}

\usepackage[numbers]{natbib}
\usepackage{amsthm}
\usepackage{amsmath,amsfonts}
\newtheorem{definition}{Definition}

\def\tsc#1{\csdef{#1}{\textsc{\lowercase{#1}}\xspace}}
\tsc{WGM}
\tsc{QE}
\tsc{EP}
\tsc{PMS}
\tsc{BEC}
\tsc{DE}

\begin{document}
\let\WriteBookmarks\relax
\def\floatpagepagefraction{1}
\def\textpagefraction{.001}
\shorttitle{SCoRE}
\shortauthors{L. Mariotti et~al.}

\title [mode = title]{SCoRE: Streamlined Corpus-based Relation Extraction using Multi-Label Contrastive Learning and Bayesian kNN}                      
\tnotemark[1]

\tnotetext[1]{We acknowledge ISCRA for awarding this project access to the LEONARDO supercomputer, owned by the EuroHPC Joint Undertaking, hosted by CINECA (Italy).}


\author[label1]{Luca Mariotti}[orcid=0009-0004-7709-5404]
\cormark[1]
\credit{Data curation, Writing – original draft, Conceptualization of this study, Methodology, Software}
\ead{luca.mariotti@unimore.it}
\author[label1]{Veronica Guidetti}[orcid=0000-0003-2233-0097]
\credit{Formal analysis, Writing – original draft, Conceptualization of this study, Methodology, Software}
\ead{veronica.guidetti@unimore.it}
\author[label1]{Federica Mandreoli}[orcid=0000-0002-8043-8787]
\credit{Supervision, Writing – original draft, Conceptualization of this study, Methodology,Formal analysis}
\ead{federica.mandreoli@unimore.it}
\cortext[cor1]{Corresponding author}

\affiliation[label1]{organization={Department of Physical, Computer and Mathematical Sciences - University of Modena and Reggio Emilia},
            addressline={via Giuseppe Campi, 213/a}, 
            city={Modena},
            postcode={41125}, 
            state={Emilia Romagna},
            country={Italy}}

\begin{abstract}
The growing demand for efficient knowledge graph (KG) enrichment leveraging external corpora has intensified interest in relation extraction (RE), particularly under low-supervision settings. 

To address the need for adaptable and noise-resilient RE solutions that integrate seamlessly with pre-trained large language models (PLMs), we introduce SCoRE, a modular and cost-effective sentence-level RE system. SCoRE enables easy PLM switching, requires no finetuning, and adapts smoothly to diverse corpora and KGs. By combining supervised contrastive learning with a Bayesian k-Nearest Neighbors (kNN) classifier for multi-label classification, it delivers robust performance despite the noisy annotations of distantly supervised corpora.

To improve RE evaluation, we propose two novel metrics: Correlation Structure Distance (CSD), measuring the alignment between learned relational patterns and KG structures, and Precision at R (P@R), assessing utility as a recommender system. We also release Wiki20d, a benchmark dataset replicating real-world RE conditions where only KG-derived annotations are available.

Experiments on five benchmarks show that SCoRE matches or surpasses state-of-the-art methods while significantly reducing energy consumption. Further analyses reveal that increasing model complexity, as seen in prior work, degrades performance, highlighting the advantages of SCoRE’s minimal design. Combining efficiency, modularity, and scalability, SCoRE stands as an optimal choice for real-world RE applications.
\end{abstract}




\maketitle

\section{Introduction}
\label{sec:intro}
Knowledge Graphs (KGs) play a crucial role in organizing and representing structured information across various domains, enhancing applications from information retrieval to complex question-answering systems \cite{KnowledgeGraphs,KGOppChall}. Enriching KGs to ensure they remain up-to-date requires leveraging external data sources, particularly textual corpora. This need has driven extensive research into relation extraction (RE) \cite{hogan2022overview,detroja2023survey}, a KG enrichment task aiming at categorizing the entailed relation between two given KG entities mentioned in the text. For instance, given the sentence \emph{``Aspirin is commonly prescribed to reduce the risk of heart attacks''} and the entity mentions \emph{``Aspirin''} and \emph{``heart attacks''}, a relation extraction system would predict the relation \textit{``prevent''} to enrich a medical KG.

A key challenge for RE approaches, based on statistical and machine learning methods, is the limited availability of high-quality annotated data. Distant supervision (DS) addresses this issue by aligning textual corpora with existing KGs, thereby automatically generating relational labels at scale \cite{mintz2009distant,bunescu2007learning,craven1999constructing}. By focusing on relations’ existence rather than relation mentions, this approach simplifies training by increasing data availability at the expense of introducing label noise, as automatic labeling may not always align with context-specific meanings in text \cite{riedel2010modeling}. As a result, recent studies have sought to mitigate DS noise with multiple-instance learning (MIL) trading sentence-level RE with entity-pair-level RE by constructing bags of sentences \cite{riedel2010modeling,jiang2016relation}. 

Given their outstanding ability to process natural language, in recent years, RE research has increasingly relied on deep learning and pre-trained language models (PLMs) \cite{hogan2022overview,detroja2023survey,10387715,SurveyRE2024Zhao}. In this context, the overarching trend is to combine sophisticated modeling strategies and fine-tuning protocols to mitigate DS noise and achieve more accurate RE models. Specifically, many of these methods use PLM fine-tuning on DS data, often incorporating entity markers or masking strategies to guide attention mechanisms and improve precision \cite{peng2020learning}. Bag-level formulations coupled with self-attention mechanisms are frequently adopted to mitigate label noise in DS  \cite{pare,REDSandT,Yin2023DistantlySR,ZhangDSREHAT}. Some approaches further enrich sentence representations with global contextual signals and structured knowledge from KGs to improve robustness in noisy settings \cite{gao2022gcek,zhou2023latent,liu2022knowledge}. However, as advocated in \cite{hu2021knowledge}, attention-based approaches are prone to degradation with increasing noise levels, as they tend to excessively concentrate on a few high-attention sentences. Other techniques aim to refine PLM representations by performing contrastive learning (CL) pretraining or using specialized training paradigms \cite{soares2019matching,peng2020learning,chen2021cil,wan2023relation}. However, while self-/semi-supervised CL frameworks are generally robust to noise, they often demand substantial quantities of relatively clean data \cite{xue2022investigating}, which are seldom available in DS scenarios. Finally, while most works operate at the bag level, some focus on sentence-based RE leveraging attention \cite{wan2022rescue,lin2023self} or CL-based finetuning \cite{li2022hiclre,sun-etal-2023-noise}.

In this work, we consider RE from a novel perspective, emphasizing solutions that are not only accurate but also readily deployable and maintainable in realistic application scenarios. To this end, we identify several key requirements for an effective RE system: (i) it should operate at the sentence level; (ii) it should address RE as a multi-label classification task; (iii) it must be cost-effective in terms of computational resources; (iv) it must be modular enough to cope with the rapidly evolving ecosystem of PLMs and allow straightforward adjustment to different corpora and KGs.

Requirement (i) arises from the observation that even if RE performance improves, full process automation remains unattainable, and some degree of external intervention will likely remain necessary for KG enrichment. Under such conditions, bag-level RE approaches are problematic because they do not enable fine-grained predictions \cite{jia2019arnor}. Instead, this granularity is necessary to explain the model's decisions and, ultimately, enable domain-expert guidance. Similarly, given the scarcity of manually curated annotated corpus in realistic domains, reliance on fully automated DS annotation becomes necessary, thus embracing a multi-label training and prediction paradigm (requirement (ii)). 
The remaining requirements stem from a thorough analysis of the key limitations in existing solutions, which often neglect scalability and the balance between performance and computational efficiency. For instance, while additional learning modules or advanced training strategies are introduced, they may not always yield significant performance gains and further obscure the model's decision-making process. Furthermore, methods that heavily depend on fine-tuning PLMs are manageable for smaller models (e.g., BERT) but become increasingly impractical for larger architectures \cite{FineTuningIsExpensive}, raising concerns about their long-term feasibility as RE solutions. Instead, a cost-effective and modular approach is essential: an approach capable of seamlessly integrating with diverse corpora, KGs, and PLMs to ensure both ready deployment and maintenance in realistic application scenarios.

Finally, we advocate that RE evaluation needs to extend beyond standard classification metrics, such as precision, recall, F1-SCoRE, and AUC. While these metrics, adapted from binary classification to the multi-label setting, typically emphasize aggregate performance across predicted relation classes, they often fail to capture multi-label performance at the per-sample level. This evaluation gap risks misrepresenting the practical utility of RE solutions, where these systems are more valuable as recommendation engines. Additionally, a thorough evaluation of the quality of learned relational patterns in relation to the underlying KG is essential to ensure their effectiveness. 

This paper proposes a viable solution that meets all the above requirements and shows its effectiveness under realistic conditions. Specifically,  we provide the following contributions to the RE research domain:
\begin{enumerate}
\item We introduce Streamlined Corpus-based Relation Extraction (SCoRE), a modular and lightweight RE tool that seamlessly adapts to various PLMs and corpus types, from high-quality human-annotated datasets to noisier DS corpora. 
In the training step, SCoRE leverages multi-label supervised CL to develop robust relation representations resilient to DS noise. Notably, it avoids fine-tuning the PLM by treating it as an informed prior for encoding head-tail pair mentions. At inference, SCoRE relies on a multi-label Bayesian kNN approach that fully exploits the hidden space structure learned by CL. This design makes SCoRE easily adaptable to new PLM releases, ensures minimal energy consumption, and streamlines model training and maintenance.
\item We propose two novel RE evaluation metrics to complement the existing focus on predictive performance: Correlation Structure Distance (CSD) and Precision at R (P@R). CSD evaluates the alignment of learned relational patterns with the underlying KG, while P@R measures the effectiveness of RE solutions as recommender systems.
\item We release Wiki20d, an extension of Wiki20m that emulates real-world RE scenarios by annotating the training set using only KG structures.
\end{enumerate}
We evaluate SCoRE against state-of-the-art models using standard and novel metrics, including environmental impact. Experiments on five benchmarks demonstrate that SCoRE matches or surpasses leading models while reducing energy consumption and improving scalability, underscoring its potential for real-world applications. 
To further validate our choice of a minimal architectural design, we assess the impact of incorporating advanced sentence and triplet processing techniques, such as dynamic full-sentence embedding, widely advocated in the literature. The results reveal that these methods heavily depend on model fine-tuning. Without fine-tuning, they not only fail to improve performance but also lead to performance degradation while increasing computational overhead.  For reproducibility, benchmarks and source code are available at \url{https://github.com/rioma96/SCoRE}.

The paper is structured as follows. Section \ref{sec:problem} formalizes sentence-based RE as a multi-label classification task. Section \ref{sec:model} details the dataset creation process, training methodology, and evaluation framework for SCoRE. In Section \ref{sec:metrics}, we describe both established and novel metrics used to benchmark our approach. The datasets employed in our experiments, including a comprehensive overview of Wiki20d, are presented in Section \ref{sec:datasets}. Section \ref{sec:experiments} outlines the state-of-the-art solutions used for comparison and the experimental setup of SCoRE. Section \ref{sec:results} provides a detailed comparison of SCoRE with state-of-the-art models, along with an analysis of the impact of modifying input, architecture, and inference configurations. Lastly, Section \ref{sec:related} discusses related work and highlights key insights, while Section \ref{sec:conclusion} draws conclusions and outlines future works.

\section{Problem Definition}
\label{sec:problem}
To formally define the problem of sentence-based multi-label RE, let us start by introducing the notion of annotated corpus and then define the task in terms of multi-label classification.
\begin{definition}[Annotated corpus]
Let us consider a reference KG denoted as $\mathcal{G} = (\mathcal{E}, \mathcal{R}, \mathcal{T})$, where $\mathcal{E}$ and $\mathcal{R}$ are the set of entities and relation types in the KG, respectively, and $\mathcal{T} \subseteq \mathcal{E} \times \mathcal{R} \times \mathcal{E}$ is the set of triples $(e_h, r, e_t)$ indicating that there exists a relation $r \in \mathcal{R}$ between the head $e_h \in \mathcal{E}$ and tail entity $e_t \in \mathcal{E}$. 

Let $\mathcal{S}=\{s_i\}_{i=1}^{\mathcal{N}}$ be a corpus of $\mathcal{N}$ sentences where $s_i\in\mathcal{S}$ denotes the $i$-th sentence and $T$ be a tokenizer which maps $s_i$ into a sequence of $T_i$ tokens $T(s_i)=[t_1,\dots, t_{T_i}]$.\\
A $\mathcal{G}$-based annotation of corpus $\mathcal{S}$ is denoted as \[\mathcal{C}^{\mathcal{G},\mathcal{S}} = \{(s_i, \{(\tau_{i,h_{j}}, \tau_{i,t_{j}}, \mathcal{R}_{i,j})\}_{j=1}^{m_i})\}_{i=1}^{\mathcal{N}}\] where:
\begin{itemize}
\item $s_i\in\mathcal{S}$ is the $i$-th sentence in $\mathcal{S}$;
\item $\tau_{i,h_{j}} (\tau_{i,t_{j}})$ denotes the set of indexes of the  tokens  in  $T(s_i)$  corresponding to the mention of the head (tail) entity of the $j$-th entity pair $(e_{h_{j}}, e_{t_{j}})\in \mathcal{E}\times \mathcal{E}$ within the sentence $s_i$;
\item $\mathcal{R}_{i,j} \subseteq \mathcal{R}$ is the set of relation types connecting  $e_{h_{j}}$  to $e_{t_{j}}$ in $s_i$. 
\end{itemize}
In the following $\mathcal{C}^{\mathcal{G},\mathcal{S}}$ will be defined as an annotated corpus.
\end{definition}
The definition above is general enough to encompass different setups. For instance, any manually annotated corpus usually assigns one relation $r\in\mathcal{R}$ to each entity pair mentions $(e_{h_{j}}, e_{t_{j}})$ and thus $\mathcal{R}_{i,j}$ is a singleton, while distantly-supervised annotated corpus usually refers to a subset of the existing relations connecting $e_{h_{j}}$ to $e_{t_{j}}$ in $\mathcal{G}$ and therefore $\mathcal{R}_{i,j}\subseteq\{r\mid (e_{h_{j}},r,e_{t_{j}})\in\mathcal{T}\}$.

\begin{definition}[Sentence-based Relation Extraction (RE)]
Given an annotated corpus $\mathcal{C}^{\mathcal{G},\mathcal{S}}$,
the goal of sentence-based relation extraction is to learn a multi-label classification function $f: \mathcal{S} \times \mathbb{N} \rightarrow 2^{\mathcal{R}}$ that, for each $s_i \in \mathcal{S}$ and $j\in[1,m_i]$, 
predicts the subset of relation types $\hat{\mathcal{R}}_{i,j}\subset\mathcal{R}$ connecting the mentions of the entity pair $(e_{h_{j}}, e_{t_{j}})$ in $s_i$.
\end{definition}
In realistic scenarios, where obtaining an annotated corpus for every domain is often unfeasible, any sentence-based RE solution must also be evaluated under real-world conditions where ground truth is unavailable during the training phase. In this case, the training algorithm can only rely on corpora annotated with information derived from the KG. This leads to the introduction of a specific kind of annotated corpus.
\begin{definition}[Fully Distantly Supervised (FDS) annotated corpus]
Let $\mathcal{C}^{\mathcal{G},\mathcal{S}}$ be annotated corpus split into a training and test set,  $\mathcal{C}_{train}^{\mathcal{G},\mathcal{S}'}$, $\mathcal{C}_{test}^{\mathcal{G},\mathcal{S}''}$, so that $\mathcal{S}'\cup \mathcal{S}''=\mathcal{S}$ and $\mathcal{S}'\cap \mathcal{S}''=\emptyset$. 
$\mathcal{C}^{\mathcal{G},\mathcal{S}}$ is \emph{fully distantly supervised} when, for each $s_i \in \mathcal{S}'$, each $R_{i,j}\in \mathcal{C}_{train}^{\mathcal{G},\mathcal{S}'}$ is the full set of relation types connecting $e_{h_{j}}$ to $e_{t_{j}}$ in $\mathcal{G}$, i.e, $R_{i,j}=\{r\mid (e_{h_{j}},r,e_{t_{j}})\in\mathcal{T}\}$.
\end{definition}

\begin{figure*}[t]
    \centering
    \includegraphics[width=1.0\textwidth]{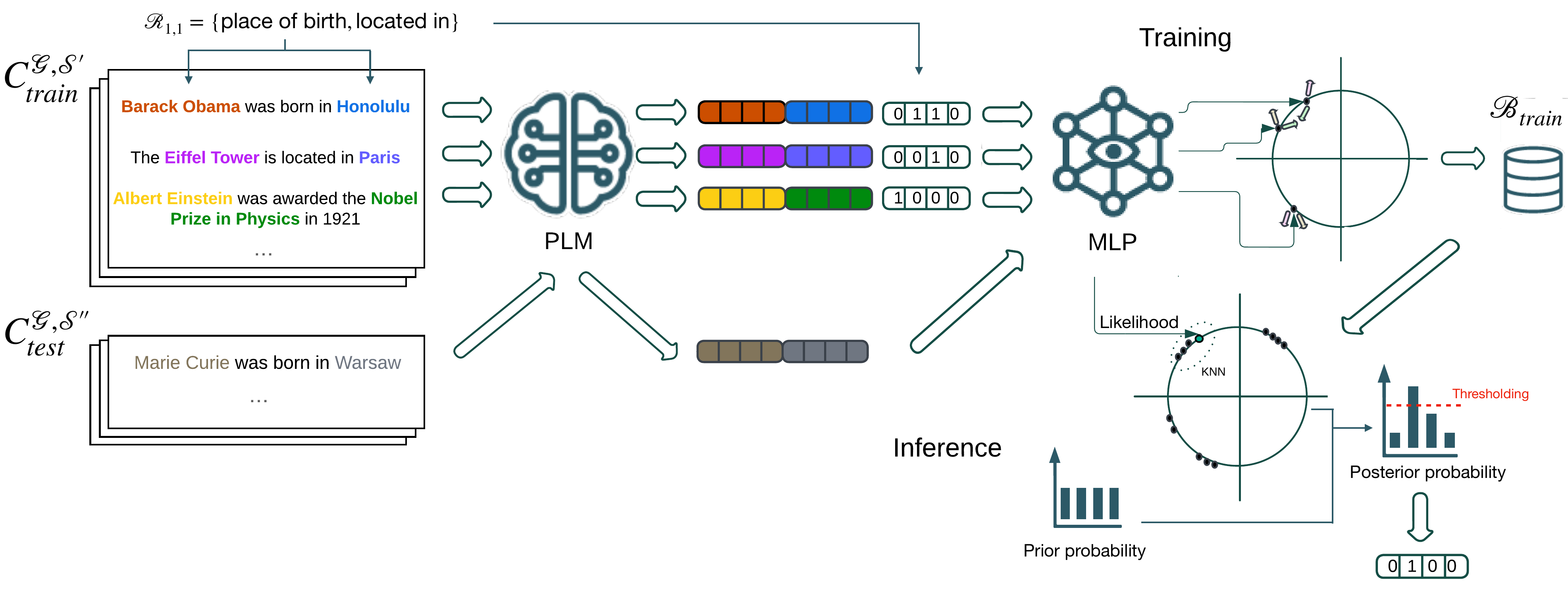} 
    \caption{Visual representation of SCoRE, showcasing the dataset creation, training, and testing stages.}
    \label{fig:SCoRE_pipeline}
\end{figure*}
\section{SCoRE}
\label{sec:model}
Figure \ref{fig:SCoRE_pipeline} illustrates the overall workflow of SCoRE. In the training phase:
\begin{enumerate} 
\item each sentence in $\mathcal{C}_{train}^{\mathcal{G},\mathcal{S}'}$ goes through a single forward pass of a PLM encoder to get an hidden vector representation of its head and tail entity mention pairs. Each hidden vector is associated with the one-hot vector of the related relation types. 
\item A multi-layer perceptron (MLP) maps head-tail encodings onto a hypersphere under a multi-label supervised CL framework, clustering samples with similar relational patterns. 
\end{enumerate} 
In the testing phase, relation type prediction on unseen head-tail entity mention pairs contained in the sentences of $\mathcal{C}_{test}^{\mathcal{G},\mathcal{S}''}$ is performed by first getting the corresponding encodings and then using a non-parametric multi-label Bayesian kNN approach. 

\subsection{Dataset Creation}
\label{sec:dataset}
Given an annotated corpus $\mathcal{C}^{\mathcal{G},\mathcal{S}}$,  each sentence $s_i\in \mathcal{C}^{\mathcal{G},\mathcal{S}}$, for $i\in[1,\mathcal{N}]$,  is mapped into a sequence of token embeddings via a single forward pass through the $\text{Encoder}$ function of the PLM: 
\begin{equation}
\mathbf{H}_i = \text{Encoder}(T(s_i)) = [\mathbf{h}_{i,1}, \mathbf{h}_{i,2}, \dots, \mathbf{h}_{i,T_i}]
\end{equation}
where $\mathbf{h}_{i,l} \in \mathbb{R}^h$ is the embedding of the $l$-th token in $s_i$.

Then, we construct the input vector $\mathbf{x}_{i,j}$ of each entity mention pair $(\tau_{i,h_{j}} \tau_{i,t_{j}})$, for $j\in [1,m_i]$, by concatenating the average embedding of the corresponding tokens:
\begin{equation}
\mathbf{x}_{i,j} = [\mathbf{e}_{i,h_{j}}; \mathbf{e}_{i,t_{j}}] \in \mathbb{R}^{2h}.
\end{equation}
where
\begin{equation}
\mathbf{e}_{i,h_j} = \frac{1}{|\tau_{i,h_j}|} \sum_{t \in \tau_{i,h_j}} \mathbf{h}_{i,t}\;,\quad \mathbf{e}_{i,t_j} = \frac{1}{|\tau_{i,t_j}|} \sum_{t \in \tau_{i,t_j}} \mathbf{h}_{i,t}.
\end{equation}
The output vector $\mathbf{y}_{i,j}$ corresponding to $\mathbf{x}_{i,j}$ is  represented as a one-hot encoding $\mathbf{y}_{i,j} \in \{0, 1\}^R$, where $R=|\mathcal{R}|$ is the number of relation types. This vector represents  set of  relation types $\mathcal{R}_{i,j}$ connecting  $e_{h_{j}}$  to $e_{t_{j}}$ in $s_i$, i.e. $y_{i,j}^k=(\mathbf{y}_{i,j})^k=1$ if $r_k \in \mathcal{R}_{i,j}$, and $y_{i,j}^k=0$ otherwise. 

The aforementioned method can be seamlessly applied to both the training $\mathcal{C}_{train}^{\mathcal{G},\mathcal{S}'}$ and test $\mathcal{C}_{test}^{\mathcal{G},\mathcal{S}''}$ corpora, giving rise to a training and test dataset.  For simplicity, we introduce a unified indexing scheme in the following, with a slight abuse of notation, and refer to both the training and test datasets as:

\begin{equation*}
\mathcal{D}_{train}=\{(\mathbf{x}_{i}, \mathbf{y}_{i})\}_{i=1,\dots,N}\, \qquad
\mathcal{D}_{test}=\{(\tilde{\mathbf{x}}_{i}, \tilde{\mathbf{y}}_{i})\}_{i=1,\dots,\tilde{N}}
\end{equation*}

where $N=\sum_{i:s_i\in\mathcal{S}'} m_i$ and $\tilde{N}=\sum_{i:s_i\in\mathcal{S}''}  m_i$ are the total number of training and test data samples respectively, with the double indexing dropped for clarity.

Note that the LLM encoder is employed solely during dataset creation, serving as an informed prior. This design prevents overfitting and reduces hallucinations that could arise from fine-tuning a large, expressive model on a limited and noisy training set, thereby preserving the encoder’s ability to interpret raw text broadly. The approach can be seamlessly adapted to any PLM. Even with models supporting larger context windows or hidden dimensions, the memory overhead increases only with the hidden representation. By storing only the averaged head and tail embeddings, the storage footprint remains limited to two token-level encodings. Moreover, as the annotated corpus requires just one PLM forward pass, the overall computational cost remains minimal.

\subsection{CL Architecture and Loss Function}
The next step consists of a supervised CL solution \cite{khosla2020supervised} relying on the following architecture. The input gets processed by a simple MLP architecture of $l$ layers with $m$ neurons, each with $l\ll m$ to ensure perturbative and stable behavior \cite{roberts2022principles}. This is followed by a single smaller layer with $m_h < m$ neurons whose outputs get normalized to unit vectors based on $L_2$ norm and thus mapped onto the hypersphere $\mathbb{S}^{m_h}$. 

To enable the prediction of multiple relations between entity pairs, we employ the adaptation of \cite{khosla2020supervised} to multi-label settings proposed by \cite{wang2022contrastive} that works as follows.
Given the training set $\mathcal{D}_{train}=\{(\mathbf{x}_{i}, \mathbf{y}_{i})\}_{i=1,\dots,N}$, a distance measure $d(\cdot , \cdot)$, and calling $\mathbf{z}_i=\text{MLP}(\mathbf{x}_i) \in \mathbb{S}^{m_h}$ the encoding of the i-th sample via the MLP, the loss function is given by: 
\begin{equation}
\label{eq:lossml}
\mathcal{L}= - \frac{1}{N}\sum_{i\in[1,N]} \sum_{j\in[1,N],j\neq i} \beta_{ij}\log\frac{e^{d(\mathbf{z}_i, \mathbf{z}_j)/\tau}}{\sum_{k\in[1,N],k\neq i} e^{d(\mathbf{z}_i, \mathbf{z}_k)/\tau}}
\end{equation}
where $\tau \in \mathbb R^+$ is a temperature parameter, and
\[ \beta_{ij}=\frac{\mathbf{y}_i^T \cdot \mathbf{y}_j}{\sum_{k\in[1,N],k\neq i} \mathbf{y}_i^T \cdot \mathbf{y}_k}\,.\]
It can be easily seen that $\mathbf{y}_i^T \cdot \mathbf{y}_j$ is the number of shared relations between the two encoded sets and $\beta_{ij}$ is its normalized version. 

This formulation can be straightforwardly used with various distance measures $d(\cdot, \cdot)$. In particular, in this work, we consider the Euclidean and cosine distances. 

The MLP weights are trained for a certain number of epochs to minimize $\mathcal{L}$. This formulation ensures that, by the end of the training, any two entity pairs that are connected through similar sets of relation types in \(\mathcal{D}_{train}\) are positioned close together in the hidden feature space. In contrast, pairs linked by distinct relationships are placed farther apart.

\subsection{Bayesian kNN for Class Prediction} 
As opposed to standard CL-based methods for classification, which follow a two-step approach \cite{khosla2020supervised}, i.e., using CL for pretraining followed by classification, we estimate test set relation type probabilities probabilities right after CL training by leveraging a probabilistic kNN approach in the hidden feature space. 

In particular, similarly to \cite{wang2022contrastive}, we create a multi-dimensional datastore of the training set hidden representations and labels $\mathcal{B}_{train}=\{\mathbf{z}_i,\mathbf{y}_i\}_{i=1\dots N}$.  Afterward, for each test example, $\tilde{\mathbf{x}}_j$, we obtain its hidden representation $\tilde{\mathbf{z}}_j = MLP(\tilde{\mathbf{x}}_j)$ and search the datastore $\mathcal{B}_{train}$ for its k-nearest neighbors $\text{kNN}(\tilde{\mathbf{z}}_j)$, according to the distance measure $d(\cdot, \cdot)$ used in the CL loss.

Then, we compute $P(r_h\mid\tilde{\mathbf{z}}_j)$, for each $r_h\in\mathcal{R}$, as follows.
As shown in \cite{gottcke2021handling}, a probabilistic multi-class kNN method requires estimating the posterior class probabilities as:
\begin{equation} 
\label{eq:pknn}
P(r_h|\tilde{\mathbf{z}}_j)= \frac{P(\tilde{\mathbf{z}}_j|r_h)\cdot P(r_h)}{P(\tilde{\mathbf{z}}_j)} \,. \end{equation}
where $P(\tilde{\mathbf{z}}_j|r_h)$ is the probability density of $\tilde{\mathbf{z}}_j$ given by the kNNs conditional on class $r_h$, $P(r_h)$ is the prior class probability, and $P(\tilde{\mathbf{z}}_j)=\sum_{i=1}^R P(\tilde{\mathbf{z}}_j|r_i)\cdot P(r_i)$ is the sample evidence. In standard situations, a common assumption is that the points are equally distributed and $ P(\tilde{\mathbf{z}}_j|r_h)\propto \frac{k_h}{n_h V(\tilde{\mathbf{z}}_j)}$; 
where $k_h$ is the number of neighbors belonging to class $r_h$ and $V(\tilde{\mathbf{z}}_j)$ is the volume of the hypersphere centered at $\tilde{\mathbf{z}}_j$ and containing all kNNs. However, CL enforces similarity between samples based on label set overlap, so a uniform density of points is not realistic. A reasonable solution is to model the point probability conditioned on each class following the CL loss function formulation as follows:
\begin{equation} 
P(\tilde{\mathbf{z}}_j|r_h) \displaystyle \propto \displaystyle \sum_{i: \mathbf{z}_i \in \text{kNN}(\tilde{\mathbf{z}}_j)} y_{i}^{h}\cdot  e^{-\frac{d(\tilde{\mathbf{z}}_j,\mathbf{z}_i)}{\tau}}\, .
\label{eq:likelihood}
\end{equation}

To make the aforementioned approach suitable for multi-label classification, we leverage Equation \ref{eq:pknn} to perform $R$ probabilistic kNN binary classification problems where the k-th task aims to predict whether the relation $r_h$ appears or not in the entity pair mention encoded in $\tilde{\mathbf{x}}_j$. In practice, the absence of a label in kNN members is treated as explicit information disfavoring the presence of such label in the final prediction. The posterior class probabilities are then given by:
\begin{equation}
P(r_h|\tilde{\mathbf{z}}_j) = \displaystyle\frac{\displaystyle\sum_{i: \mathbf{z}_i \in \text{kNN}(\tilde{\mathbf{z}}_j)} P(r_h) \cdot y_{i}^{h} \cdot e^{-\frac{d(\tilde{\mathbf{z}}_j, \mathbf{z}_i)}{\tau}}}{\displaystyle \sum_{i: \mathbf{z}_{i} \in \text{kNN}(\tilde{\mathbf{z}}_j)}\left[P(r_h)\cdot y_{i}^{h} + P(\overline{r_h}) \bar{y}_{i}^{h}\right] e^{-\frac{d(\tilde{\mathbf{z}}_j, \mathbf{z}_{i})}{\tau}}}   
\end{equation}
where $P(\overline{r_h})=(1- P(r_h))$ is the probability that the relation $r_h$ does not appear in the kNN labels (so that $P(r_h)+P(\overline{r_h})=1$) and $\bar{y}_{i}^{h}=(1- y_{i}^{h})$.
Selecting an appropriate prior probability $P(r_h)$ is crucial in Bayesian inference as it reflects initial beliefs about class distributions before observing the data. The choice of prior can significantly influence predictions, especially in imbalanced or small datasets where data-driven likelihoods may be insufficient to reliably infer class probabilities.

In our setting, we follow the prescription of \cite{gottcke2021handling} and use an uninformative or flat prior distribution for each binary classification problem, i.e., the prior probabilities that a class appears or does not appear in the kNNs are equal, i.e., $P(r_h)=1/2 \; \forall h\in[1,R]$. Choosing flat prior probabilities is equivalent to adopting a dynamic, local neighborhood-dependent class weight. This strategy focuses on optimizing recall over precision, thereby minimizing the risk of missing positive instances (false negatives) \cite{gottcke2021handling}. In imbalanced classification problems, such as classifying relation type distributions in text, prioritizing recall is essential as it improves the detection of long-tail relation types (true positives), even if it leads to an increase in false positives. Moreover, if CL training is successful, relation types will be far from equally distributed in the hidden feature space, making the flat prior assumption even more reasonable. 

After computing posterior class probabilities, we use the following universal thresholding to get sharp class predictions $\hat{\mathbf{y}}(\tilde{\mathbf{x}}_j)$:
\begin{equation}
(\hat{\mathbf{y}}(\tilde{\mathbf{x}}_j))^h=\hat{y}_j^h= 
\begin{cases}
1, & \text{if } P(r_h|\tilde{\mathbf{z}}_j) > c, \\
0, & \text{otherwise}.
\end{cases}
\label{eq:thresholding}
\end{equation}
where $c\in[0,1]$ is a threshold commonly shared among classes.

\section{Evaluation Metrics}
\label{sec:metrics}
In this section, we outline the evaluation criteria for assessing both the performance and environmental impact of sentence-level RE solutions. We start with commonly used RE metrics, explicitly defining their extensions to multi-label classification. Moreover, we illustrate the process we use to rank prediction by model confidence and define the process used to measure environmental impact. Finally, we introduce additional metrics not yet explored in RE literature that are essential for evaluating result quality and reliability of results. 
\subsection{MicroF1 and MacroF1}
Two of the most common metrics for performance evaluation in RE are micro- and macro-averaged $F_1$ SCoRE, named microF1 and macroF1, respectively. While originally used in multi-class classification models, these metrics can be straightforwardly adapted to multi-label settings. 
In particular, 
\begin{equation}
\text{microF1} = \frac{TP}{TP + 0.5 \cdot (FN + FP)}
\end{equation}
where $TP$, $FP$, and $FN$ denote the total number of true positives, false positives, and false negatives across all classes respectively. This metric emphasizes the overall accuracy of the model in predicting the correct relationships for all instances.
The macroF1 SCoRE is obtained by computing the microF1 SCoRE for each class individually and then averaging the results:
\begin{equation}
\text{macroF1} = \frac{1}{R} \sum_{h=1}^{R} \frac{TP_h}{TP_h + 0.5 \cdot (FN_h + FP_h)}\,.
\end{equation}
This formula helps evaluate the model's performance across different relation types, regardless of their frequency in the dataset.
Note that in those datasets where only part of the labels appear in the test set, we adopt the convention of considering in macroF1 only those classes for which $FP_h+FN_h+TP_h>0$ as the class-specific $F_1$-SCoRE would be unspecified otherwise.

\subsection{Ranking Solutions by Model Confidence: Micro- and MacroF1@M}
A standard metric applied in RE problems consists of reporting the microF1 and macroF1 values (or the precision values) on the M test set elements for which the model is most confident, denoted as microF1@M and macroF1@M, respectively. While ranking solutions according to model confidence is straightforward in multi-class classification\footnote{it corresponds to sorting the test set elements according to the values of the maximum class posterior probability}, 
in multi-label setting the exact formula to be employed was not clearly defined in most works. 

In this paper, we apply the following scoring function to rank test set elements by model confidence. Calling $P_{\tilde{\mathbf{z}}_j}=\{P(r_h|\tilde{\mathbf{z}}_j)_{i=1,\dots, R}\mid \hat{y}_j^h=1\}$ and $|P_{\tilde{\mathbf{z}}_j}|$ the cardinality of this set $s(\tilde{\mathbf{z}}_j)$, we define the confidence SCoRE of the model on the test sample $\tilde{\mathbf{z}}_j$ as the harmonic mean of the posterior class probabilities associated with positive predictions:
\begin{equation}
s(\tilde{\mathbf{z}}_j)=\left[\prod_{P(r_h|\tilde{\mathbf{z}}_j)\in P_{\tilde{\mathbf{z}}_j}} P(r_h|\tilde{\mathbf{z}}_j) \right]^{1/|P_{\tilde{\mathbf{z}}_j}|}\,.
\label{eq:ranking}
\end{equation}
The reason behind this choice is the following. If we consider the presence of different classes as independent events, then the probability of observing multiple classes is simply given by the product of the individual class probabilities. However, choosing this product as the final SCoRE would disproportionately penalize predictions containing multiple labels. By using the harmonic mean we create a standard SCoRE that removes the distortion associated with multiple labels summarizing them in a single "probability value". Once the data are ranked by confidence $s(\tilde{\mathbf{z}}_j)$, micro-F1 and macro-F1 SCoREs can be recomputed on the top M most confident samples.

We note that, while @M metrics can be useful for evaluating model calibration under idealized conditions, they are influenced by the number of classes present in the most confident samples, a quantity typically unknown a priori. Consequently, these metrics often emphasize performance on a subset of relation classes, potentially distorting overall performance assessments.

\subsection{Carbon Footprint}
Given the critical importance of sustainability in real-world settings, we extended our evaluation to include an energy consumption metric, aiming to support the development of resource-efficient relation classification methods. Energy consumption was quantified in kilowatt-hours (kWh) using CodeCarbon \cite{codecarboon}, a well-established tool that automatically detects the hardware specifications and measures consumption in real-time based on system usage.

\subsection{P@R: Precision at R}
P@R is commonly used in information retrieval to evaluate the relevance of a system’s top R ranked predictions. In sentence-level RE, it provides an instance-level indicator of how effectively a model performs as a recommender, where ranking quality is paramount and domain experts primarily judge the correctness of the top few results. Formally, P@R is the proportion of true positives among the top R predictions, reflecting the model’s ability to pinpoint relevant classes for each instance. When R matches the number of ground-truth relevant classes per instance, P@R effectively measures the model’s precision at exactly the point where all true relevant labels should appear.

To compute this metric, given the number true labels $R_j = \sum_{i\in[1,R]}\tilde y_j^i $ for each test sample $\tilde{\mathbf{x}}_j$, we rank the posterior probabilities $P(r_h|\tilde{\mathbf{z}}_j)$ in increasing order and create a vector $\hat{\mathbf{y}}_{R_j}(\tilde{\mathbf{x}}_j)$ assigning a positive prediction to the top $R_j$ posterior probabilities. 

The P@R SCoRE is then defined as the average value of the ratio of true positives in the top $R$ posterior probabilities for all test samples:
\begin{equation}
\text{P@R} = \frac{1}{\tilde{N}}\sum_{j=1}^{\tilde{N}}\frac{|\hat{\mathbf{y}}_{R_j}(\tilde{\mathbf{x}}_j) \cap \tilde{\mathbf{y}}_j|}{R_j}.
\end{equation}
This approach offers a direct evaluation of the model’s ability to retrieve all relevant labels without considering irrelevant predictions, as long as the number of relevant classes aligns with R. Furthermore, it adapts to instances with varying numbers of true relation classes. 

\subsection{Correlation Structure Distance (CSD)}
In multi-label RE, the relational types that co-occur for the same entity mention pair, frequently exhibit non-trivial relationships. For instance, the relation ``lives in'' is often accompanied by ``born in'' for a person–location pair, while relations such as ``father of'' and ``mother of'' (or ``son of'') for a person–person pair should never appear together. 

Traditional performance metrics do not quantify how well a model preserves the underlying correlation structure among labels. To achieve this goal, we introduce the Correlation Structure Distance (CSD), which quantifies the discrepancy between the correlation matrix of the true test relation types and that of the predicted ones, providing a deeper evaluation of model robustness and alignment with the underlying data structure.  Specifically, using a set of one-hot encoding vectors $\mathbf{Y}=\{\mathbf{y}\}_{j=1\dots,\tilde{N}}$, we compute the correlation between each relation pair $(r_h,r_p)$, for each $r_h,r_p\in\mathcal{R}$ using the Pearson $\phi$ coefficient:
\begin{equation}
\phi(\mathbf{Y},r_h,r_p)=\frac{n_{11}^{(h,p)} n_{00}^{(h,p)} - n_{01}^{(h,p)} n_{10}^{(h,p)}}{\sqrt{n_{1\cdot}^{(h,p)} n_{0\cdot}^{(h,p)} n_{\cdot 1}^{(h,p)} n_{\cdot 0}^{(h,p)} }}
\end{equation}
where \begin{itemize}
\item $n_{11}^{(h,p)}=\sum_{j=1}^{\tilde{N}}y_j^h  y_j^p$  and $n_{00}^{(h,p)}=\sum_{j=1}^{\tilde{N}}(1- y_j^h)(1- y_j^p)$ are the number of co-occurrences of positive and negative labels in the two classes, respectively; 
\item $n_{01}^{(h,p)}=\sum_{j=1}^{\tilde{N}}(1- y_j^h) y_j^p$ and $n_{10}^{(h,p)}=\sum_{j=1}^{\tilde{N}}\hat y_j^h(1- y_j^p)$ count the number of cases where the labels disagree; 
\item $n_{1\cdot}^{(h,p)} = \sum_{j=1}^{\tilde{N}} y_j^h$ and $n_{\cdot1}^{(h,p)}=\sum_{j=1}^{\tilde{N}}\hat y_j^p$ indicate the total number of cases where $r_h$ or $r_p$ get value 1 in $\hat{\mathbf{Y}}$ respectively. Similar considerations hold for $n_{0\cdot}$ and $n_{\cdot 0}$. 
\end{itemize}
Given the true test set labels $\tilde{\mathbf{Y}}=\{\tilde{\mathbf{y}}_j\}_{j=1\dots,\tilde{N}}$ and the predicted ones $\hat{\mathbf{Y}}=\{\hat{\mathbf{y}}(\tilde{x}_j)\}_{j=1\dots,\tilde{N}}$, we compute the distance between $\phi(\tilde{\mathbf{Y}},r_h,r_p)$ and $\phi(\hat{\mathbf{Y}},r_h,r_p)$ using the Frobenius norm: 
\begin{equation}
\text{CSD} = \sqrt{\sum_{h=1}^R \sum_{p=1}^R \left|\phi(\hat{\mathbf{Y}},r_h,r_p)-\phi(\tilde{\mathbf{Y}},r_h,r_p)\right|^2 }\,.
\end{equation}
The CSD values are inversely proportional to the ability of the model to reflect the ground truth relationships between labels.

\begin{table*}[t]
    \centering
    \caption{Summary of datasets and their characteristics after pre-processing operations.}
    \begin{tabular}{|c|c|c|c|c|c|c|c|c|c|}
        \hline
        \textbf{Dataset} & \textbf{Ref. KG} & \textbf{Ann. Train} & \textbf{Ann. Test} & \textbf{N. Rel} & \textbf{Len Train} & \textbf{Len Val} & \textbf{Len Test} & \textbf{ML Train } &  \textbf{ML Test } \\
        \hline
        \textbf{DisRex-Eng}    & Wikidata  & DS & DS   & 36   & 128,241  & 18,304   & 33,399 & 18\% & 27\%\\
        \hline
        \textbf{Wiki20m}   & Wikidata  & DS & Man  & 80   & 276,260  & 17,485   & 101,861 &3\% & 1\%\\
        \hline
        \textbf{NYT10M}    & Freebase  & DS & Man  & 24   & 72,961   & 9,172    & 6,642 & 12\%&  19\%\\
        \hline
        \textbf{NYT10D}    & Freebase  & DS & DS   & 55   & 80,542  & --       & 4,892 &  44\%&  21\%\\
        \hline
        \textbf{Wiki20D}   & Wikidata  & FDS & Man  & 755  & 614,207   & 56,187   & 92,083 & 20\% & 1\% \\
        \hline
    \end{tabular}
    \label{sec:datasets_table}
\end{table*}

\section{Datasets}
\label{sec:datasets}
Five annotated corpus were considered to conduct the experiments: four are well-established benchmarks for RE, while the fifth, Wiki20d, is our released FDS annotated corpus. Their features are summarized in Table \ref{sec:datasets_table}, detailing for each annotated corpus the reference KG, the generation method (DS - Distant Supervised, Man - Manual, FSD - Full Distant Supevised),  the total number of relation types, and dataset split size. The table also highlights the varying complexities of these datasets in terms of the number of relations and the percentage of multi-label instances in their training and test sets (ML Train/Test).

Specifically, the first four benchmarks are: \textbf{NYT10d} \cite{riedel2010modeling}, a benchmark constructed via DS by linking mentions in the New York Times corpus to Freebase; \textbf{NYT10m}\cite{manualdatasets}, a manually curated version of NYT10d with enhanced annotation quality and additional validation splits; \textbf{Wiki20m}\cite{manualdatasets} generated via DS by aligning Wikipedia articles with Wikidata and providing a manually annotated test set; \textbf{DisRex}\cite{disrex}, a multilingual dataset created using DS and Wikidata, designed to balance relation types and include inverse relations to verify that a model truly learns the proper ordering of entity pairs. Since we use the English version of BERT, we consider only the English portion of DisRex.

\begin{figure}[t]
    \centering
    \includegraphics[width=\columnwidth]{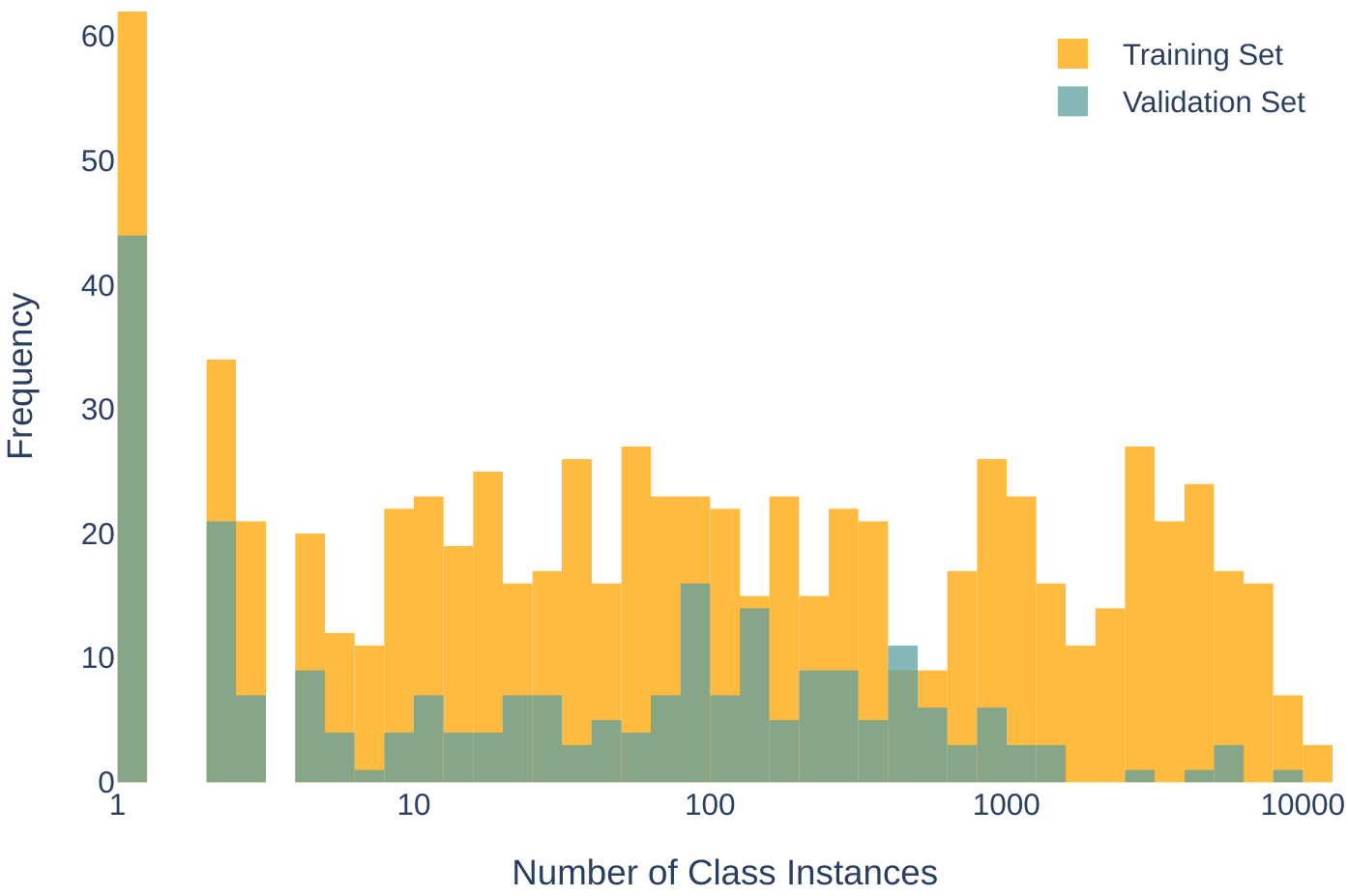}
    \caption{Wiki20D relation class distribution.}
    \label{fig:wiki20dclassdistribution}
\end{figure}

Finally, we introduce \textbf{Wiki20d}, an extension of Wiki20m obtained by processing each sentence and querying Wikidata to extract relations between identified head-tail entity pairs. By relying on FDS, Wiki20d features a substantially larger number of relation classes that are heavily imbalanced (see Fig \ref{fig:wiki20dclassdistribution}), enabling a more realistic assessment of methods in scenarios that demand the extraction of complex, multi-label relational information.

To ensure consistent preprocessing across datasets, and align them with the multi-label RE task, we applied several data-cleaning steps. First, we removed sentences labeled with the 'NA' relation, as they introduced no meaningful relationships and could distort performance assessments. We also removed sentences where entity token positions exceeded the PLM context window, as they would impede proper representation learning. 

\section{Experimental Setup}
\label{sec:experiments}

\subsection{Settings Used in SOTA Models}
We compare SCoRE with four sentence-level RE methods and one bag-level RE method:
\begin{itemize}
    \item \textbf{KGPool} \cite{kgpool}: a sentence-level RE approach that employs a dynamic context augmentation mechanism to incorporate only the KG facts directly relevant to the sentence. The method utilizes a BiLSTM to independently encode representations for sentences, entities, and the contextual information of KG entities, constructing a Heterogeneous Information Graph (HIG). To refine this structure, KGPool applies a Self-Attention-Based Graph Convolutional Network to reduce the HIG into a more concise Context Graph. Finally, a context aggregator is employed to jointly learn from both the Context Graph and sentence-level representations.
    \item \textbf{PARE} \cite{pare}: a bag level RE approach  that concatenates all sentences within a bag—defined by the head and tail entities—into a single passage by sequentially sampling sentences without replacement. The passage is then encoded using BERT to generate contextualized embeddings for every token. A relation query vector is subsequently employed to create a relation-aware summary of the entire passage through an attention mechanism. This summary is fed into an MLP, which outputs the probability of the corresponding relation triple.
    \item \textbf{HiCLRE} \cite{li2022hiclre}: a sentence and bag level approach employing multi-granularity contextualization to capture cross-level structural information using multi-head self-attention across entity, sentence, and bag levels. This recontextualization aligns context-aware features from each level, refining semantic representations. Additionally, dynamic gradient adversarial perturbation improves robustness by using gradient-based CL to create pseudo-positive samples, enhancing the model's ability to distinguish relations. 
    \item \textbf{SSLRE} \cite{sun-etal-2023-noise}: a sentence-level approach that addresses the noise in DS by discarding only the labels of noisy samples and treating these instances as unlabeled. It employs a weighted k-NN graph to select confident samples as labeled data, while the rest are treated as unlabeled. The framework then uses a semi-supervised learning approach to handle remaining label noise and effectively utilize unlabeled samples.
    \item \textbf{TIW} \cite{lin2023self}: a sentence-level RE tool that uses a transitive instance weighting mechanism combined with self-distilled BERT for denoise DS sentence-level training in RE. The method fine-tunes the BERT encoder, then fixes its parameters and trains student classifiers using knowledge distillation. TIW generates dynamic instance weights to reduce noise and overfitting by considering uncertainty and consistency. Students choose between the teacher's and previous peer's outputs based on consistency, while false negative filtering and positive weighting adjust weights for negative and positive instances, respectively.
\end{itemize}
Since SCoRE operates at the sentence level, its results can be directly compared with all competitors except PARE without modifications. However, we were only partially able to reproduce KGPool’s results due to the limited flexibility of the code, which made adapting it to different datasets beyond those used in the original paper challenging. Additionally, because the codes for \cite{sun-etal-2023-noise} and \cite{lin2023self}  were not publicly released, we could only report the results presented in their original publications.

In contrast, since PARE is designed for bag-level predictions, we adapted it for sentence-level evaluation. To achieve this, we trained the model at the bag level and, during the prediction phase, provided one sentence at a time as input. This approach guarantees that PARE predictions are directly comparable to sentence-level methods. PARE's flexible design enabled us to reproduce results for all the datasets considered.

\subsection{SCoRE Configuration}
\label{sec:configuration}
For dataset creation, we employed the BERT-base model, utilizing a context window of 512 tokens, an embedding dimension of 768, and 12 transformer layers. We performed a grid search across a range of hyperparameters to optimize the MLP model and the Bayesian kNN approach. The grid search was conducted on all datasets, with the following parameter configurations:

\begin{itemize}
    \item \texttt{MLP - layers }($l$): [3,4,\textbf{5}]
    \item \texttt{MLP - depth/width }($l/m$): [\textbf{0.01}, 0.05, 0.1]
    \item \texttt{MLP - output dims }$(\text{m}_h)$: [5, 10, \textbf{15}]
    \item \texttt{MLP - activation}: [\textbf{\texttt{swish}}, ReLU]
    \item \texttt{loss - distance}: [\textbf{\texttt{euclidean}}, \texttt{cosine}]
    \item \texttt{loss - temperature}$(\tau)$: [\textbf{0.01}, 0.05, 0.1, 0.2]
    \item \texttt{learning\_rate}: [$10^{-4}$, $\mathbf{5\cdot 10^{-3}}$, $10^{-3}$]
    \item \texttt{batch\_size}: [64, 128, \textbf{256}]
    \item \texttt{kNN} ($k$): [5, 10, 15, 50, 100, 150]
    \item \texttt{probability threshold }$(c)$: [0.3, 0.4, 0.5, 0.6, 0.7]
\end{itemize}

Each experiment was conducted over 30 training epochs, using the AdamW optimizer, with early stopping employed to monitor the validation set loss, when available. Early stopping was triggered when the loss ceased to decrease, with a patience of 5 epochs, and the model weights were reverted to their best configuration at the point of optimal performance. In particular, we identified a common configuration for the CL training that performed well across all datasets (highlighted in bold). The only hyper-parameters that could not be fixed across datasets are related to the prediction stage: $c$ and \texttt{kNN}. 
This difference is reasonable, as \texttt{kNN} regulates the local probability density estimate used to compute posterior class probabilities, while the probability threshold, $c$, determines how to convert these probabilities into sharp predictions. The optimal values for these hyperparameters may depend on dataset-specific characteristics, such as differences in dataset balance and the nature of the underlying relations. The optimal threshold and kNN values for each dataset were selected based on results from the validation set, or from the training set in the absence of a validation set. The corresponding values are the following:
\begin{itemize}
    \item \texttt{NYT10M}: $c$: 0.6 and $k$: 50
    \item \texttt{NYT10D}: $c$: 0.7 and $k$: 100
    \item \texttt{DISREX}: $c$: 0.5 and $k$: 50
    \item \texttt{WIKI20M}: $c$: 0.5 and $k$: 100
    \item \texttt{WIKI20D}: $c$: 0.7 and $k$: 150
\end{itemize}

\section{Results}
\label{sec:results}

\begin{table*}[t]
    \centering
    \caption{Performance comparison of SCoRE with state-of-the-art models using micro-F1 and macro-F1 metrics. Results from non-reproducible methods marked by an asterisk (*)}
        \begin{tabular}{|c|c|c|c|c|c|c|}
            \hline
            \textbf{Model} & \textbf{Metric} & \textbf{NYT10M} & \textbf{NYT10D} & \textbf{DisRex} & \textbf{Wiki20m} & \textbf{Wiki20D} \\
            \hline
            \multirow{2}{*}{KGPOOL} & microF1 & - & 72.0 & -    & -    & - \\
                                    & macroF1 & - & 41.8 & -    & -    & - \\
            \hline
            \multirow{2}{*}{HiCLRE} & microF1 & 68.7 & 75.2  & 61.2 & \textbf{87.6} & \textbf{68.3} \\
                                  & macroF1   & 34.7 & 18.4  & 51.3 & \textbf{86.3} & 06.9 \\
            \hline
            \multirow{2}{*}{SSLRE} & microF1  & $63.8^*$ & -  & - & $81.5^*$ & - \\
                                    & macroF1 &  -       & -  & - & -       & - \\
            \hline
            \multirow{2}{*}{TIW}    & microF1  & $63.8^*$ & $55.3^*$  & - & - & - \\
                                    & macroF1 &  $35.2^*$ & -         & - & \underline{$84.1^*$} & - \\
            \hline
            \multirow{2}{*}{PARE} & microF1 & \underline{77.5} & \underline{86.4}  & \textbf{77.1} &       83.8 & 67.2 \\
                                  & macroF1 & \underline{38.9} & \underline{46.3}  & \textbf{70.5} & 83.7 & \underline{19.6} \\
            \hline
            \multirow{2}{*}{SCoRE} & microF1 & \textbf{77.6} & \textbf{89.2}  & \underline{75.4} & 83.5 & \underline{66.9} \\
                                     & macroF1 & \textbf{40.8} & \textbf{49.6}  & \underline{62.6} & 80.7 & \textbf{23.9} \\
            \hline
        \end{tabular}

    \label{tab:competitorsf1}
\end{table*}
All experiments were conducted on the Leonardo Booster partition, using a BullSequana X2135 “Da Vinci” single-node GPU Blade equipped with a 32-core Intel Xeon Platinum 8358 CPU (Ice Lake), 512 GB of DDR4 RAM, and four NVIDIA Ampere A100 GPUs (64GB HBM2e) interconnected via NVLink 3.0. 

\subsection{Performance Against SOTA}
In this section, we provide a detailed evaluation of our model's performance and carbon footprint, comparing them against state-of-the-art models. 

\noindent
\textbf{Micro and MacroF1.}
We begin by comparing the models' performance using the commonly adopted micro- and macro-F1 metrics across various benchmark datasets. As previously mentioned, we were only partially able to reproduce some results due to limited code flexibility or the absence of publicly available repositories. The results in Table \ref{tab:competitorsf1} present the best performance over five runs replicating the competitors.

Our findings indicate that SCoRE consistently delivers performance that is competitive with or superior to other approaches while maintaining a minimal architecture. Specifically, SCoRE outperforms competitors on the NYT10D, NYT10M, and Wiki20d datasets. Indeed, although HiCLRE achieves the highest micro-F1 on Wiki20d, its macro-F1 SCoRE is significantly lower. On the Wiki20m dataset, HiCLRE attains the highest performance; however, results on other datasets suggest that this approach is better suited for datasets with low multi-label complexity and balanced class distributions. A similar pattern is observed for TIW, although the lack of reproducibility limits our ability to draw definitive conclusions. PARE achieves the second-highest micro-F1 on Wiki20m, slightly surpassing SCoRE. On the DisRex dataset, PARE outperforms SCoRE by a few percentage points in both micro- and macro-F1 SCoREs.

These results demonstrate that relying on sophisticated architectures or complex training structures, such as pooling knowledge graph information to enrich embeddings (KGPool), does not always lead to improved performance and may hinder adaptation to diverse contexts. Additionally, extensive engineering efforts can reduce a RE method’s adaptability to various dataset characteristics, such as multi-label complexity and class imbalance, as seen with HiCLRE. Nonetheless, PARE, a model that integrates complex contextual embeddings through attention mechanisms and fine-tunes the PLM, performs well across datasets, with results only marginally higher or lower than SCoRE. Therefore, we focus our further analyses on comparing our solution with PARE. Notably, PARE's higher micro- and macro-F1 SCoREs on DisRex suggest that complex contextual embeddings via attention mechanisms may be more effective for such dataset. To further investigate this hypothesis, we now examine the label correlation matrix distance using the CSD.

\begin{figure}[t]
    \centering
    \includegraphics[width=\columnwidth]{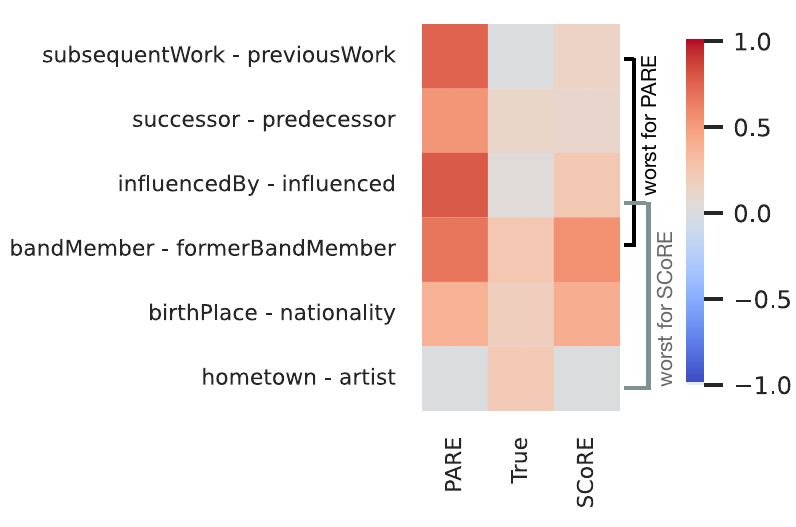}
    \caption{PARE and SCoRE worst correlation distance entries.}
    \label{fig:cdsworstpareSCoRE}
\end{figure}

\noindent
\textbf{Label Correlation Structure Distance (CSD).}
The comparison between SCoRE and PARE CSD values in Table~\ref{fig:cdsworstpareSCoRE} highlights that SCoRE consistently outperforms PARE across all five benchmark datasets, with lower CSD values signifying better alignment with the true relational structure. This is especially notable for the NYT10M, NYT10D, and Wiki20D datasets, which feature highly imbalanced relation labels and substantial DS noise, underscoring SCoRE's effectiveness in maintaining accurate relational alignments despite these challenges. Also in the Wiki20M dataset, which contains only a small percentage of multi-label samples in both the training (3\%) and test sets (1\%), SCoRE still demonstrates superior alignment.

\begin{table}[t]
   \centering
   \caption{Label correlation matrix distance via the CSD.}
   \begin{tabular}{|c|c|c|}
        \hline
       \textbf{Dataset} & \textbf{PARE} & \textbf{SCoRE} \\
       \hline
       \textbf{NYT10M} & 1.38 & \textbf{1.08} \\
       \hline
       \textbf{NYT10D} & 1.23 & \textbf{0.43} \\
       \hline
       \textbf{DisRex} & 1.99 & \textbf{1.15} \\
       \hline
       \textbf{Wiki20m} & 0.90 & \textbf{0.18} \\
       \hline
       \textbf{Wiki20D} & 5.30 & \textbf{2.31} \\
       \hline
    \end{tabular}
   \label{tab:competitorsdistances}
\end{table}

Surprisingly, despite PARE's strong F1 performance, its CSD values are much higher than SCoRE's on the DisRex dataset. This counterintuitive result is further clarified through the analysis presented in Figure \ref{fig:cdsworstpareSCoRE}, which compares the highest 4 misalignment of the correlation matrix entries between PARE and SCoRE against the ground truth labels. The analysis reveals that PARE struggles to differentiate between relations with opposite meanings, such as ``followed'' versus ``followed-by'' and ``previous work'' versus ``subsequent work.'' This limitation is critical in RE tasks, as it can lead to the conversion of asymmetric relations into symmetric ones, thereby distorting the logical structure of the KG. Furthermore, this misalignment may violate transitivity properties, resulting in erroneous inferences and compromising the integrity of the KG.

\begin{table}[t]
   \centering
   \caption{Comparison of Energy Consumption (kWh)}
   \begin{tabular}{|c|c|c|c|c|}
        \hline
        \textbf{Dataset} & \makecell{\textbf{PARE}\\ \textbf{SL}} & \makecell{\textbf{PARE}\\\textbf{BAG}} & \multicolumn{2}{c|}{\textbf{SCoRE}} \\
        \cline{4-5}
                          &                 &                  & \textbf{Dataset} & \textbf{Train \& Test} \\
        \hline
        \textbf{NYT10M}   & 14.78           & 2.11             & 0.63                      & 0.002                     \\
        \hline
        \textbf{NYT10D}   & 12.39           & 1.47             & 0.81                      & 0.003                     \\
        \hline
        \textbf{DisRex}   & 271.67          & 3.93             & 1.43                      & 0.003                     \\
        \hline
        \textbf{Wiki20m}  & 743.25          & 2.55             & 2.07                      & 0.005                     \\
        \hline
        \textbf{Wiki20D}  & 1561.84         & 6.20             & 5.75                      & 0.06                      \\
        \hline
   \end{tabular}
   \label{tab:competitorsenergy}
\end{table}

\noindent
\textbf{Environmental Impact.}
We now compare the energy consumption of our solution, SCoRE, with that of PARE using CodeCarbon, which measures energy usage in kilowatt-hours (kWh). Specifically, for SCoRE, energy consumption was recorded separately during dataset preprocessing and the training/testing phases. In contrast, for PARE, a single measurement encompassed the entire process, as its model dynamically generates embeddings during each training and testing cycle. The results, presented in Table \ref{tab:competitorsenergy}, demonstrate significant differences in energy efficiency, further stressed by the distinct data processing methods employed by each model.

PARE was adapted to operate on a sentence-by-sentence basis, resulting in considerable energy overhead. In this configuration, PARE creates a virtual test dataset containing only one sentence, necessitating repeated dataset loading, embedding calculation, and inference for each individual sentence during testing. This inefficiency leads to high energy consumption, particularly for larger datasets such as Wiki20m (743 kWh). Conversely, the SCoRE model separates dataset creation from training and testing, thereby significantly reducing energy usage. For instance, SCoRE's dataset creation phase for Wiki20d requires 5.75 kWh, while subsequent training and testing demand only minimal energy (0.06 kWh for Wiki20d). For smaller datasets, SCoRE's energy consumption during training and testing is as low as 0.002–0.003 kWh, offering a more sustainable and energy-efficient approach, especially for large-scale datasets where PARE's inefficiency is most pronounced.

Additionally, it is important to note that even when PARE is utilized for bag-level RE, its energy consumption for training and testing remains substantially higher than that of SCoRE. This is due to the necessity of repeated embedding calculations and LLM weight updates for each training epoch. For example, training and testing PARE on Wiki20m at the bag level consumes approximately 2.5 kWh. In realistic settings, where training or continual learning must be performed periodically to keep the system updated, adopting a RE solution that matches the efficiency of existing models while reducing energy consumption by up to three orders of magnitude significantly enhances sustainability and lowers operational costs.

\subsection{Anti-Ablation Analysis: The Adverse Effects of Architectural Enhancements.}
\begin{table}[t]
    \centering
    \caption{Effects of Architectural Enhancements.}
    \begin{tabular}{|c|c|c|c|c|c|c|}
        \hline
        \textbf{Dataset} & \textbf{Metric} & \textbf{SCoRE + CLS} & \textbf{SE+A} & \textbf{SE+A+CL} \\
        \hline
        \multirow{3}{*}{NYT10M} & microF1 & 77.4 & 65.1 & 72.5 \\
                                & macroF1 & 40.4 & 23.2 & 23.7 \\
                                & kWh     & 0.64 & 3.88 & 7.8 \\
        \hline
        \multirow{3}{*}{NYT10D} & microF1 & 89.8 & 67.4 & 75.1 \\
                                & macroF1 & 48.9 & 23.3 & 23.9 \\
                                & kWh     & 0.92 & 14.42 & 15.12 \\
        \hline
        \multirow{3}{*}{DisRex} & microF1 & 75.2 & 36.2 & 52.7 \\
                                & macroF1 & 63.1 & 23.6 & 31.3 \\
                                & kWh     & 1.44 & 4.05 & 18.15 \\
        \hline
        \multirow{3}{*}{Wiki20m} & microF1 & 83.8 & 63.7 & 63.6 \\
                                 & macroF1 & 81.2 & 58.1 & 59.9 \\
                                 & kWh     & 2.08 & 13.54 & 19.5 \\
        \hline
        \multirow{3}{*}{Wiki20d} & microF1 & 66.8 & 48.6 & 50.6 \\
                                 & macroF1 & 22.9 & 15.6 & 16.4 \\
                                 & kWh     & 5.83 & 40.1 & 57.3 \\
        \hline
    \end{tabular}
    \label{tab:ablationf1}
\end{table}
In contrast to ablation studies, which identify critical model components by systematically removing modules and measuring the resulting performance decline, we investigate the impact of increasing architectural complexity. Specifically, the experiments in this section are designed with the following objectives. The overall results are presented in Table \ref{tab:ablationf1}. 
\paragraph{\textbf{SCoRE + CLS}} In the experiment, we assess whether the input structure associated with encoding head-tail pairs during dataset creation (Section \ref{sec:dataset}) is sufficient or if it can benefit from additional information, without substantially modifying the SCoRE pipeline. To this aim we explored the impact of incorporating the CLS token into triplet mention embeddings, following the standard approach of concatenating the CLS embedding with those of the head and tail entities appearing in the sentence \cite{wan2023relation}. A grid-search was performed to find the best hyperparameters, following Section \ref{sec:configuration}. The results in Table \ref{tab:ablationf1}, reveal that, contrary to expectations, the inclusion of the CLS token did not lead to any performance variation. Both micro and macro F1 SCoREs remained largely unchanged, suggesting that the CLS token does not provide substantial contributions to the entity pair embeddings. \\

Subsequently, considering the robust results of PARE across datasets of various types, we investigate whether replacing the dataset creation phase with a more commonly used approach in the literature can lead to improved outcomes. To this end, we perform two experiments. The first is not directly related to the SCoRE architecture, it represents a sentence-level adaptation of PARE without fine-tuning and serves solely as a comparison for the final experiment. The second, replaces SCoRE's dataset creation phase with PARE's dynamic sentence encoding mechanism and attention mechanism.
\paragraph{\textbf{Sentence Embedding + Attention (SE+A)}} In this experiment, we explore whether PARE's approach, which involves dynamic sentence encoding supported by an attention mechanism for RE, can remain effective when the fine-tuning of the PLM is removed and training happens at the sentence level. Freezing PLM weights requires additional modifications to the input processing compared to PARE. In fact, the use of attention mechanisms in RE tasks typically yields suboptimal performance unless entity markers or masks are employed \cite{peng2020learning, chen2021cil}. While PARE uses special entity markers, this approach is not feasible in our case, as the model cannot learn new special tokens without fine-tuning. Consequently, we adopt the strategy of using MASK tokens to replace the head and tail mentions and train PARE's architecture using a multi-label cross-entropy loss and the hyperparameter setting described in \cite{pare}. We train the model for 60 epochs and use early stopping as described in Section \ref{sec:configuration}. The results for the SE+A configuration demonstrate significant underperformance relative to both SCoRE and PARE, underscoring the critical role of PLM fine-tuning in attention-based models that leverage full sentence embeddings.
\paragraph{\textbf{Sentence Embedding + Attention + CL (SE+A+CL)}}
This final experiment involves utilizing PARE's dynamic sentence encoding mechanism and attention mechanism by replacing SCoRE's dataset creation phase. Therefore, in this version, sentences are processed dynamically and follow PARE's architecture with masked head and tail entity as in SE+A. However, the remainder of the learning and inference process adheres to the steps defined in the SCoRE pipeline: supervised multi-label CL based on the attention head's output and Bayesian kNN during inference. To make this possible, we removed the softmax activation function from the last layer of SE+A and added the minimal component to perform CL sensibly, i.e. a single fully connected layer with a small number of neurons $m_h$ whose output get normalized to unit vectors based on $L_2$ norm. This model is then trained using the multilabel CL loss function described in Equation \ref{eq:lossml}. A grid-search was performed to find the best hyperparameters for the loss, learning rate, batch size, $m_h$, kNN, and probability threshold, as described in Section \ref{sec:configuration}. The results show that, although the SE+A+CL approach underperforms SCoRE, it yields comparable or even superior results relative to the SE+A configuration. This indicates the effectiveness of supervised CL in managing noisy datasets, as it emphasizes learning robust representations that differentiate between similar and dissimilar instances.

Overall, these experiments demonstrate that increasing input expressiveness or modifying the model’s architecture can often degrade performance, particularly when PLM fine-tuning is avoided. This underSCoREs that the interplay between complexity and performance is not always a straightforward trade-off.

\begin{table*}[ht]
\begin{small}
    \centering
    \caption{Performance on overall datasets (left) and most confident M samples (right) across Bayesian kNN configurations.}
    \begin{tabular}{|c|c|cccc|cccc|}
        \hline
        \multirow{2}{*}{\textbf{Dataset}}& \multirow{2}{*}{\textbf{kNN}}  & \multicolumn{4}{c|}{\textbf{Overall performance}}  & \multicolumn{4}{c|}{\textbf{@M performance}}\\
        \cline{3 - 10}
         & & \textbf{microF1} & \textbf{macroF1} & \textbf{CSD} & \textbf{P@R} & \textbf{m@100} & \textbf{M@100} & \textbf{m@1000} & \textbf{M@1000}\\
        \hline
        \multirow{4}{*}{NYT10D} 
        & UU & \underline{89.1 $\pm$ 0.5} & \underline{48.8 $\pm$ 0.6} & \textbf{0.4 $\pm$ 0.0} & \textbf{92.0 $\pm$ 0.4} &\underline{93.3 $\pm$ 0.4} & 68.2 $\pm$ 6.1 & \underline{93.2 $\pm$ 0.4} & \underline{53.1 $\pm$ 0.9}\\
        & UC & 78.7 $\pm$ 0.5 & 28.2 $\pm$ 0.7 & 7.0 $\pm$ 0.0 & \textbf{92.0 $\pm$ 0.4} & \textbf{94.7 $\pm$ 1.1} & \textbf{95.2 $\pm$ 0.6} & \textbf{95.2 $\pm$ 0.6} & \textbf{59.8 $\pm$ 3.9}  \\
        & IU & 86.9 $\pm$ 0.3 & 40.8 $\pm$ 0.6 & \underline{0.7 $\pm$ 0.1} & \underline{91.3 $\pm$ 0.6} & 92.9 $\pm$ 1.7 & \underline{72.5 $\pm$ 2.0} & 90.8 $\pm$ 0.8 & 52.0 $\pm$ 0.8  \\
        & IC & \textbf{89.2 $\pm$ 0.6} & \textbf{48.8 $\pm$ 1.0} & 0.8 $\pm$ 0.1 & \underline{91.3 $\pm$ 0.6} & 92.9 $\pm$ 2.8 & 71.4 $\pm$ 15.4 & 93.1 $\pm$ 0.5 & 52.0 $\pm$ 6.0  \\
        \hline
        \multirow{4}{*}{NYT10M} 
        & UU  & \textbf{77.5 $\pm$ 0.1} & 40.0 $\pm$ 0.9 & \textbf{1.1 $\pm$ 0.0} & \underline{78.4 $\pm$ 0.3} & \underline{86.6 $\pm$ 4.0} & \underline{60.7 $\pm$ 13.7} & \underline{85.9 $\pm$ 0.2} & \underline{43.5 $\pm$ 1.6} \\	 
        & UC & 69.0 $\pm$ 0.4 & \textbf{42.7 $\pm$ 0.6} & 3.0 $\pm$ 0.1 & \underline{78.4 $\pm$ 0.3} & 85.9 $\pm$ 0.8 & 46.9 $\pm$ 2.0  & \textbf{86.6 $\pm$ 1.0} & 42.0 $\pm$ 1.4 \\
        & IU  & 76.8 $\pm$ 0.7 & 32.4 $\pm$ 1.1 & \textbf{1.1 $\pm$ 0.0} & \textbf{80.6 $\pm$ 0.3} & \textbf{87.8 $\pm$ 4.2} & 54.0 $\pm$ 4.8 & 85.4 $\pm$ 0.3 & 42.9 $\pm$ 1.9 \\
        & IC & \underline{77.1 $\pm$ 0.2} & \underline{40.5 $\pm$ 0.8} & \underline{1.2 $\pm$ 0.0} & \textbf{80.6 $\pm$ 0.3} & 86.6 $\pm$ 0.7 & \textbf{63.5 $\pm$ 5.8} & 85.6 $\pm$ 0.2 & \textbf{43.7 $\pm$ 4.0} \\
        \hline
        \multirow{4}{*}{DISREX} 
        & UU  & \textbf{75.2 $\pm$ 0.2} & \textbf{62.9 $\pm$ 0.5} & \textbf{1.2 $\pm$ 0.1} & \textbf{76.7 $\pm$ 0.1} & \textbf{86.6 $\pm$ 1.9} & \underline{64.2 $\pm$ 7.6} & \textbf{86.7 $\pm$ 0.3} & \underline{58.2 $\pm$ 1.8} \\
        & UC & 60.8 $\pm$ 0.3 & \underline{51.9 $\pm$ 0.5} & 4.4 $\pm$ 0.0 & \textbf{76.7 $\pm$ 0.1}  & 83.8 $\pm$ 1.9 & \textbf{66.2 $\pm$ 4.7} & 85.9 $\pm$ 0.2 & \textbf{66.6 $\pm$ 4.0} \\
        & IU  & \underline{65.3 $\pm$ 0.1} & 38.0 $\pm$ 1.0 & \underline{1.7 $\pm$ 0.1} & \underline{76.5 $\pm$ 0.2} & 84.4 $\pm$ 2.3 & 55.8 $\pm$ 5.5 & 84.0 $\pm$ 1.0 & 56.1 $\pm$ 3.6 \\
        & IC & \textbf{75.2 $\pm$ 0.2} & \textbf{62.9 $\pm$ 0.5} & \textbf{1.2 $\pm$ 0.1} & \underline{76.5 $\pm$ 0.2} & \underline{85.3 $\pm$ 1.5} & 62.2 $\pm$ 3.8 & \underline{86.3 $\pm$ 0.1} & 57.5 $\pm$ 0.8 \\
        \hline
        \multirow{4}{*}{WIKI20M} 
        & UU  & \textbf{83.5 $\pm$ 0.1} & \textbf{80.8 $\pm$ 0.1} & \textbf{0.2 $\pm$ 0.0} & \textbf{83.0 $\pm$ 0.1} & 97.7 $\pm$ 1.3 & 94.5 $\pm$ 2.9 & \textbf{98.3 $\pm$ 0.1} & \textbf{93.7 $\pm$ 1.8} \\
        & UC & 62.4 $\pm$ 1.0 & \underline{65.8 $\pm$ 0.8} & 6.4 $\pm$ 0.1 & \textbf{83.0 $\pm$ 0.1} & \textbf{98.0 $\pm$ 1.4} & \textbf{95.3 $\pm$ 3.4} & 97.8 $\pm$ 0.2 & 92.1 $\pm$ 2.5 \\	 
        & IU  & \underline{70.3 $\pm$ 1.2} & 60.3 $\pm$ 1.5 & \underline{0.6 $\pm$ 0.0} & \underline{82.4 $\pm$ 0.1} & 96.7 $\pm$ 1.3 & 90.5 $\pm$ 3.8 & 98.1 $\pm$ 0.3 & \underline{93.2 $\pm$ 0.8} \\
        & IC & \textbf{83.5 $\pm$ 0.1} & \textbf{80.8 $\pm$ 0.1} & \textbf{0.2 $\pm$ 0.0} & \underline{82.4 $\pm$ 0.1} & \underline{98.0 $\pm$ 0.0} & \underline{94.6 $\pm$ 0.3} & \underline{98.1 $\pm$ 0.4} & 92.8 $\pm$ 1.6 \\
        \hline
        \multirow{4}{*}{WIKI20D}
        & UU  & \underline{65.1 $\pm$ 2.5} & \textbf{23.2 $\pm$ 0.8} & \underline{2.3 $\pm$ 0.1} & \textbf{67.2 $\pm$ 1.3} & \textbf{96.8 $\pm$ 2.4} & \underline{89.1 $\pm$ 6.9} & \textbf{95.7 $\pm$ 0.3} & 77.3 $\pm$ 3.7 \\
        & UC & 20.8 $\pm$ 1.4 & 5.1 $\pm$ 0.2 & 32.3 $\pm$ 0.4 & \textbf{67.2 $\pm$ 1.3} & 92.0 $\pm$ 2.8 & 89.0 $\pm$ 2.8 & 93.3 $\pm$ 1.2 & \underline{77.7 $\pm$ 2.7} \\
        & IU  & 24.5 $\pm$ 3.0 & 16.3 $\pm$ 1.1 & \textbf{1.0 $\pm$ 0.0} & \underline{64.7 $\pm$ 1.7} & \underline{95.7 $\pm$ 1.3} & \textbf{90.5 $\pm$ 2.4} & \underline{95.1 $\pm$ 0.2} & 75.7 $\pm$ 4.7 \\
        & IC & \textbf{67.2 $\pm$ 1.5} & \underline{21.2 $\pm$ 0.8} & 3.6 $\pm$ 0.1 & \underline{64.7 $\pm$ 1.7} & 93.7 $\pm$ 3.3 & 88.3 $\pm$ 6.6 & 95.0 $\pm$ 1.1 & \textbf{82.3 $\pm$ 0.7} \\
        \hline
    \end{tabular}
    \label{tab:dataset_comparison_tot}
\end{small}
\end{table*}

\subsection{Impact of Bayesian kNN Configurations on SCoRE Performance}
While the previous section highlighted the drawbacks of some architectural modifications, here we focus on examining whether different configurations in the model's prediction stage can enhance performance. 

Using a flat prior in Bayesian kNN neglects class frequencies, potentially overemphasizing rare classes and biasing predictions, thereby reducing accuracy for majority classes. Therefore, it may be interesting to investigate how different prior choices affect prediction outcomes. We performed a sensitivity analysis to evaluate the impact of using an informative prior, calculated from class frequencies as $P(r_i)=n_i/\sum_{j} n_j$, where $n_i$ is the number of training instances in class $r_i$. While this method can address class imbalance and improve calibration, in a kNN framework it may diminish the influence of rare classes in the posterior. Specifically, when a rare class is among the nearest neighbors, the frequency-based prior reduces its effect on the posterior probabilities.

Selecting appropriate decision thresholds is essential for ensuring result quality. Universal thresholding (Equation \ref{eq:thresholding}) applies a single threshold across all classes, while class-specific thresholding assigns distinct thresholds to each class. The latter is particularly effective in imbalanced datasets, as it helps prevent the under-prediction of rare classes by aligning sensitivity and specificity with class prevalence. However, determining optimal thresholds poses a challenge. Various methods exist, including Bayesian decision theory, cross-validation-based probability calibration, and utilizing prior class probabilities. To maintain simplicity and due to the absence of a preferred error type, we adopt the use of prior class probabilities thesholding, defined as
\begin{equation*}
\hat{y}(\tilde{\mathbf{z}})_i = 
\begin{cases}
1, & \text{if } P(r_i|\tilde{\mathbf{z}}) > P(r_i), \\
0, & \text{otherwise}.
\end{cases}
\end{equation*}

In our experiments, we tested how performance were affected by both prior and decision threshold choices. In particular, we will refer to the configurations as:
\begin{itemize}
\item{UU: Uninformative prior/Universal threshold (default SCoRE setup).}
\item{UC: Uninformative prior/Class-specific threshold.}
\item{IU: Informative prior/Universal threshold.}
\item{IC: Informative prior/Class-specific threshold.}
\end{itemize}

For each of the previous configurations, we computed micro-/macroF1 SCoREs and CSD values. Each experiment was repeated 10 times to facilitate the comparison among different configurations. For what concerns universal thresholding, we retained the threshold value providing the best performance on the validation set, where available. 

The results on the left hand side of Table \ref{tab:dataset_comparison_tot} demonstrate that the UU (SCoRE) and IC configurations consistently achieve the best micro- and macro-F1 SCoREs compared to other methods. However, UU exhibits greater agreement with the ground truth label correlation matrix, as indicated by the smaller CSD values. In contrast, methods like UC and IU experience significant performance degradation, with IU underperforming due to its sensitivity to class variability from the informative prior, and UC struggling because of its reliance on prior probability thresholds, which may not accurately reflect local label distributions. 

\subsection{From @M Performance to P@R}
Let us now focus on the reliability degree of model confidence by looking at the  m@M and M@M performance. The results, shown on the right hand side of Table \ref{tab:dataset_comparison_tot}, clearly confirm that, regardless of the configuration, all setups drastically improve performance when requested to judge only the most confident samples based on our ranking method in Equation \ref{eq:ranking}. The configuration UU confirms to be the best with 16 podium positions (6 best, 10 second best) followed by UC (9 best, 1 second best), IC (3 best, 5 second best), and finally IU (2 best, 4 second best). On this task, the UC configuration works surprisingly well, especially when compared to the relatively low performance on the whole datasets, suggesting that class-specific thresholding can provide marginal benefits. However, upon careful consideration, no single setup consistently outperforms the others across all scenarios. This is particularly surprising, especially considering the performance gap observed on the full dataset, raising questions about the actual utility of these metrics.

To more effectively evaluate model performance under real-world conditions, we shift the focus from class-based to instance-based performance by analyzing the behavior of P@R metrics across different Bayesian kNN configurations. The results, presented in Table \ref{tab:dataset_comparison_tot}, demonstrate that P@R values are consistent within each prior type and exhibit minimal variation between universal and class-specific thresholding methods. This indicates that fluctuations in micro- and macro-F1 SCoREs are largely driven by threshold selection rather than changes in the predictive model itself. Indeed, although the UU (SCoRE) and IC methods use different priors, a different choice of decision thresholds can significantly realign their predictive performance. Notably, P@R values for SCoRE are comparable to or exceed those of micro- and macro-F1 SCoREs, highlighting its strong performance as a recommender system. Furthermore, uninformative priors generally yield higher P@R values across most datasets, establishing SCoRE (Section \ref{sec:model}) as the optimal configuration among those evaluated.

\section{Related Works}
\label{sec:related}
Recent RE research has increasingly focused on deep learning solutions. Among those, non-PLM approaches often leverage alternative architectures and external knowledge resources, such as KgPool \cite{kgpool}, which enriches input context with KG facts for single-label extraction, and graph NN-based methods like RECON \cite{recon}, which align extracted relations with KG entries for improved accuracy. However such approaches can hardly be adapted to different case studies and KGs. 

In PLM-based DS RE, fine-tuning has become a dominant approach, enabling models to specialize their embeddings by adjusting parameters to capture relational patterns \cite{Szep2024APG}. 

Several studies frame RE as a bag-level classification task via MIL to address noise in DS annotations. Apart from PARE \cite{pare}, the approaches leveraging attention mechanisms aim to enhance robustness by using sophisticated learning techniques such as employing syntactic trimming \cite{REDSandT}, adding sequential layers like Bi-LSTM \cite{Yin2023DistantlySR}, or introducing hierarchical attention to align sentence- and bag-level representations \cite{ZhangDSREHAT}. 
To further mitigate noise and improve generalization, some MIL works enrich instance representations with external knowledge. For example, \cite{liu2022knowledge} integrates fine-grained alignment and inductive signals from KG neighbors to address long-tail relations. Another example \cite{gao2022gcek} combines global context with knowledge-aware embeddings to guide denoising, while the authors in \cite{zhou2023latent} reconstruct latent structural graphs and refine them through iterative optimization, leveraging pretrained KBs to improve sentence-level representation learning.
Another popular approach finetunes the PLM using CL to align sentences with similar entity pair mentions or triplets. For example, \cite{wan2023relation} uses sentence-level predictions and scores to compute prior weights to guide bag-level CL training, whereas \cite{chen2021cil} employs contrastive pre-training with sentence- and attention-derived bag encodings.

In this paper, we advocate that sentence-level RE is the only approach enabling domain expert guidance and oversight. In principle, models trained at bag-level can be used to infer single sentences, as we did with PARE. However, MIL training often focuses on high-attention sentences, under-utilizing data, and increasing noise sensitivity \cite{chen2021empower, hu2021knowledge}. Additionally, its effectiveness relies on multiple sentences per entity pair, which is challenging for long-tail relations. 
These limitations become evident in the comparison with PARE provided in Section \ref{sec:results}, where we show that it consistently predicts opposite relations (a problem that could only be highlighted thanks to sentence-level predictions). 

Some recent efforts continue to rely on PLM fine-tuning but operate at the sentence level. Among these, we directly compared SCoRE with TIW \cite{lin2023self}, HiCLRE \cite{li2022hiclre}, and SSLRE \cite{sun-etal-2023-noise} in \ref{sec:results}.
Another relevant method is \cite{wan2022rescue}, which aims to improve implicit and long-tail relation performance by combining a fine-tuned LLM's output with a memory-based kNN classifier leveraging training triplets encoding. Although Equation \ref{eq:likelihood} is similar to the one used in \cite{wan2022rescue}, their kNN formulation is heuristic, as it does not derive distance weighting from a metric induced in the hidden feature space. In contrast, SCoRE enforces the metric in the hidden feature space via CL for its kNN formulation. Furthermore, the approach in \cite{wan2022rescue} lacks a Bayesian framework, does not support multi-label prediction, and performs inference by merging the outputs of the classifier and kNN.

\section{Conclusions and Future Works}
\label{sec:conclusion}
In this paper, we introduced SCoRE, a sentence-based multi-label RE tool designed to effectively handle the noise inherent in DS annotations. To the best of our knowledge SCoRE is the first approach that completely avoids finetuning, using the PLM solely as an informed prior during dataset creation. This approach minimizes computational costs and ensures adaptability to advancements in LLMs, as the PLM is employed only in a single forward pass. Although relying solely on supervised CL may appear simplistic, this method has been demonstrated to enhance robustness against input noise and hyperparameter variability \cite{khosla2020supervised}, efficiently utilizing the information in positive pairs while retaining the capacity of CL to mine negative samples. Departing from prevailing trends in the literature, SCoRE does not use CL as a pretraining step followed by a classifier layer. Instead, we streamline the training and testing process by employing a local non-parametric method, i.e., Bayesian kNN, for direct inference on test set labels based on the metric induced by CL in the hidden feature space. Furthermore, we choose prior class probabilities to enhance recall, improving the detection of long-tail relation types and reinforcing SCoRE's capability to handle complex, imbalanced datasets. 

We demonstrated that SCoRE's minimal architecture matches or surpasses state-of-the-art models, offering a lightweight, adaptable, and interpretable solution for RE. This effectiveness is confirmed on our realistic dataset, Wiki20d, which simulates real-world conditions requiring reliance on FDS annotations. By introducing the CSD metric, we showed that SCoRE better aligns with relational patterns found in KG relations. Additionally, we highlighted the detrimental impact of incorporating more complex input and sentence-processing techniques, which, while suitable for fine-tuning approaches, proved counterproductive for SCoRE's design. Additionally, we evaluated SCoRE’s sensitivity to prior probabilities and thresholding, identifying a flat prior and universal thresholding as the optimal configuration. This analysis further reveals that micro-F1@M and macro-F1@M metrics are poor indicators of real-world performance because they primarily capture a subset of high-confidence samples, limiting their practical utility. For this reason, we recommend focusing on metrics like P@R for instance-based evaluation and a more accurate assessment of RE systems in decision support contexts. SCoRE’s strong P@R results highlight its effectiveness as a recommender system, maintaining robust performance across different prior probability settings and demonstrating substantial potential for real-world applications.

Given the promising results achieved by the SCoRE pipeline, we plan to extend our approach in future work to encompass entity linking and relational triple extraction \cite{10387715, SurveyRE2024Zhao, bidirectionTripleZhang, SimpleTripleExtrRen, StudentTeacherTripleExtractionZhao}. Additionally, we aim to enhance the applicability of SCoRE as a recommender system by integrating it into a human-in-the-loop framework \cite{HITL, CleanGraph}. This integration would facilitate expert interaction with the data-driven learning pipeline, improving model adaptability and maintenance while streamlining continuous updates.

\printcredits

\bibliographystyle{cas-model2-names}

\bibliography{SCoRE}

\end{document}